\documentclass[twocolumn,letterpaper, 10 pt, journal, twoside]{IEEEtran}
\usepackage[T1]{fontenc}
\usepackage[latin9]{inputenc}
\usepackage{color}
\usepackage{units}
\usepackage{mathrsfs}
\usepackage{amsmath}
\usepackage{amssymb}
\usepackage{graphicx}
\usepackage[unicode=true,
 bookmarks=true,bookmarksnumbered=true,bookmarksopen=true,bookmarksopenlevel=1,
 breaklinks=false,pdfborder={0 0 0},pdfborderstyle={},backref=false,colorlinks=false]
 {hyperref}
\hypersetup{pdftitle={Your Title},
 pdfauthor={Your Name},
 pdfpagelayout=OneColumn, pdfnewwindow=true, pdfstartview=XYZ, plainpages=false}

\makeatletter

\let\SF@@footnote\footnote
\def\footnote{\ifx\protect\@typeset@protect
    \expandafter\SF@@footnote
  \else
    \expandafter\SF@gobble@opt
  \fi
}
\expandafter\def\csname SF@gobble@opt \endcsname{\@ifnextchar[
  \SF@gobble@twobracket
  \@gobble
}
\edef\SF@gobble@opt{\noexpand\protect
  \expandafter\noexpand\csname SF@gobble@opt \endcsname}
\def\SF@gobble@twobracket[#1]#2{}
\providecommand{\tabularnewline}{\\}

\let\oldforeign@language\foreign@language
\DeclareRobustCommand{\foreign@language}[1]{%
  \lowercase{\oldforeign@language{#1}}}

\usepackage[caption=false,font=footnotesize]{subfig}

\usepackage{graphicx}
\usepackage{import}

\usepackage{balance}

\usepackage{resizegather}

\makeatother

\begin{document}
\global\long\def\quat#1{\boldsymbol{#1}}%

\global\long\def\dq#1{\underline{\boldsymbol{#1}}}%

\global\long\def\hp{\mathbb{H}_{p}}%

\global\long\def\dotmul#1#2{\langle#1,#2\rangle}%

\global\long\def\partialfrac#1#2{\frac{\partial\left(#1\right)}{\partial#2}}%

\global\long\def\totalderivative#1#2{\frac{d}{d#2}\left(#1\right)}%

\global\long\def\mymatrix#1{\boldsymbol{#1}}%

\global\long\def\vecthree#1{\operatorname{vec}_{3}#1}%

\global\long\def\vecfour#1{\operatorname{vec}_{4}#1}%

\global\long\def\haminuseight#1{\overset{-}{\mymatrix H}_{8}\left(#1\right)}%

\global\long\def\hapluseight#1{\overset{+}{\mymatrix H}_{8}\left(#1\right)}%

\global\long\def\haminus#1{\overset{-}{\mymatrix H}_{4}\left(#1\right)}%

\global\long\def\haplus#1{\overset{+}{\mymatrix H}_{4}\left(#1\right)}%

\global\long\def\norm#1{\left\Vert #1\right\Vert }%

\global\long\def\abs#1{\left|#1\right|}%

\global\long\def\conj#1{#1^{*}}%

\global\long\def\veceight#1{\operatorname{vec}_{8}#1}%

\global\long\def\myvec#1{\boldsymbol{#1}}%

\global\long\def\imi{\hat{\imath}}%

\global\long\def\imj{\hat{\jmath}}%

\global\long\def\imk{\hat{k}}%

\global\long\def\dual{\varepsilon}%

\global\long\def\getp#1{\operatorname{\mathcal{P}}\left(#1\right)}%

\global\long\def\getpdot#1{\operatorname{\dot{\mathcal{P}}}\left(#1\right)}%

\global\long\def\getd#1{\operatorname{\mathcal{D}}\left(#1\right)}%

\global\long\def\getddot#1{\operatorname{\dot{\mathcal{D}}}\left(#1\right)}%

\global\long\def\real#1{\operatorname{\mathrm{Re}}\left(#1\right)}%

\global\long\def\imag#1{\operatorname{\mathrm{Im}}\left(#1\right)}%

\global\long\def\spin{\text{Spin}(3)}%

\global\long\def\spinr{\text{Spin}(3){\ltimes}\mathbb{R}^{3}}%

\global\long\def\distance#1#2#3{d_{#1,\mathrm{#2}}^{#3}}%

\global\long\def\distancejacobian#1#2#3{\boldsymbol{J}_{#1,#2}^{#3}}%

\global\long\def\distancegain#1#2#3{\eta_{#1,#2}^{#3}}%

\global\long\def\distanceerror#1#2#3{\tilde{d}_{#1,#2}^{#3}}%

\global\long\def\dotdistance#1#2#3{\dot{d}_{#1,#2}^{#3}}%

\global\long\def\distanceorigin#1{d_{#1}}%

\global\long\def\dotdistanceorigin#1{\dot{d}_{#1}}%

\global\long\def\squaredistance#1#2#3{D_{#1,#2}^{#3}}%

\global\long\def\dotsquaredistance#1#2#3{\dot{D}_{#1,#2}^{#3}}%

\global\long\def\squaredistanceerror#1#2#3{\tilde{D}_{#1,#2}^{#3}}%

\global\long\def\squaredistanceorigin#1{D_{#1}}%

\global\long\def\dotsquaredistanceorigin#1{\dot{D}_{#1}}%

\global\long\def\crossmatrix#1{\overline{\mymatrix S}\left(#1\right)}%

\global\long\def\constraint#1#2#3{\mathcal{C}_{\mathrm{#1},\mathrm{#2}}^{\mathrm{#3}}}%

\title{Virtual Fixture Assistance for Suturing in Robot-Aided Pediatric Endoscopic
Surgery}
\author{Murilo~M.~Marinho$^{*}$,~\IEEEmembership{Member,~IEEE,} Hisashi~Ishida$^{*}$,
Kanako~Harada,~\IEEEmembership{Member,~IEEE,} Kyoichi~Deie, and
Mamoru~Mitsuishi,~\IEEEmembership{Member,~IEEE}\thanks{$^{*}$The
authors made equal contributions.}\thanks{Manuscript received: September,
9, 2019; Revised November, 30, 2019; Accepted December, 24, 2019.}\thanks{This
paper was recommended for publication by Editor Pietro Valdastri upon
evaluation of the Associate Editor and Reviewers' comments. This work
was supported by JSPS KAKENHI Grant Number 19H05585. \emph{(Corresponding
author:} Murilo M. Marinho.)}\thanks{Murilo M. Marinho, Hisashi
Ishida, Kanako Harada, and Mamoru Mitsuishi are with the Department
of Mechanical Engineering, the University of Tokyo, Tokyo, Japan.
\texttt{Emails:\{murilo, h.ishida, kanako, mamoru\}@nml.t.u-tokyo.ac.jp}.
}\thanks{Kyoichi Deie is with the Department of Pediatric Surgery,
Kitasato University Hospital, Kanagawa, Japan. \texttt{Email:kdeie@med.kitasato-u.ac.jp}}\thanks{Digital
Object Identifier (DOI): 10.1109/LRA.2019.2963642.}}
\markboth{IEEE Robotics and Automation Letters. Preprint Version. Accepted December,
2019}{Murilo M. Marinho \MakeLowercase{\emph{et al.}}: Virtual Fixture
Assistance for Suturing}
\maketitle
\begin{abstract}
The limited workspace in pediatric endoscopic surgery makes surgical
suturing one of the most difficult tasks. During suturing, surgeons
have to prevent collisions between instruments and also collisions
with the surrounding tissues. Surgical robots have been shown to be
effective in adult laparoscopy, but assistance for suturing in constrained
workspaces has not been yet fully explored. In this letter, we propose
guidance virtual fixtures to enhance the performance and the safety
of suturing while generating the required task constraints using constrained
optimization and Cartesian force feedback. We propose two guidance
methods: looping virtual fixtures and a trajectory guidance cylinder,
that are based on dynamic geometric elements. In simulations and experiments
with a physical robot,\textcolor{black}{{} we show that the proposed
methods increase precision and safety }\textcolor{black}{\emph{in-vitro}}\textcolor{black}{.}
\end{abstract}

\begin{IEEEkeywords}
Medical Robots and Systems, Kinematics, Collision Avoidance
\end{IEEEkeywords}

\section{Introduction}

\setlength{\textfloatsep}{0pt}
\setlength{\intextsep}{10pt}
\setlength{\abovecaptionskip}{0pt}

\IEEEPARstart{P}{ediatric endoscopic surgery} for infants (<1 year
old) and neonates has additional difficulties when compared with adult
endoscopic surgery. \textcolor{black}{For example, the workspace is
narrower, often being described by medical doctors as \textquotedblleft golf-ball
sized\textquotedblright . Constrained workspaces can be defined as
having a volume of 200~cm$^{3}$ or less \cite{Cundy2015a}, which
includes pediatric and neonatal patients.} The limited workspace increases
the risks of collisions between instruments, which can occur both
inside and outside the patient. These difficulties have motivated
the usage of surgical robots, such as the da Vinci Surgical System
(Intuitive Surgical Inc., USA), which have been shown to improve dexterity,
endurance, and vision. The da Vinci Surgical System has had great
success in adult laparoscopy; however, it has been shown to be inapplicable
to pediatric surgery owing to the large diameter of its instruments
(8 mm) and the required in-patient length (5 cm) \cite{Takazawa2018}.

To provide appropriate robotic assistance to pediatric surgery and
other applications in constrained workspaces, our group has been developing
a novel master-slave robotic system, called SmartArm, in parallel
with this work \cite{marinho2019integration}. It consists of a pair
of industrial robot arms, each of which is instrumented with an actuated
flexible instrument \cite{Arata2019}. The proposed system has instruments
whose diameters are 3.5 mm, and the preliminary results indicate that
our system can operate inside constrained workspaces, such as those
in pediatric patients \cite{Marinho2019a}. With the SmartArm system,
we expect to bridge the gaps that prevent the wide adoption of robots
in pediatric and neonate surgery \cite{Cundy2015}.

As in other surgical robotics scenarios, robotic assistance in pediatric
surgery may increase the task completion time owing to motion scaling
as reported in the literature \cite{Osa2018}, especially when performing
complex tasks. Suturing is among the most complex surgical procedures.
It requires bimanual manipulation of a needle, a thread, and the target
tissue. Suturing can be divided into four steps: (1) the instrument
grabs the needle, (2) the needle is inserted through both sides of
the tissue, (3) one of the instruments grabs the thread near the needle
and loops the thread a few times around the other instrument, and
finally (4) the loose end of the thread is pulled to tighten the knot.
To compensate for possible robot assistance drawbacks and further
improve task performance, many groups have proposed assistance methods
for suturing subtasks or a combination of subtasks \cite{kapoor2005spatial,kapoor2006constrained,Sen2016,chen2016virtual,Osa2018,Fontanelli2018}.

One of those methodologies is task automation \cite{Sen2016,Osa2018},
which has been so far demonstrated in an unobstructed space, which
is not the case in pediatric endoscopy. Moreover, although the future
potential of such techniques is clear, currently they are still outperformed
by human-operated robots and are unable to leverage surgical skills
efficiently.

In contrast, virtual fixtures do not aim to fully automate the task.
Instead, virtual fixtures are used to enhance the operator's medical
skills. A comprehensive survey on virtual fixtures was presented by
Bowyer \emph{et al.} \cite{bowyer2014active}. The survey shows that
virtual fixtures are often built using geometric elements such as
points, lines, planes, and corresponding volumetric primitives. They
are divided into \emph{regional virtual fixtures}, to create a forbidden
region or safe zone, and \emph{guidance virtual fixtures}, to aid
the operator in performing specific tasks.

In this letter, we focus on the generation of guidance virtual fixtures
for the looping stage of suturing in an endoscopic pediatric surgery
setting. Looping can be time-consuming and requires considerable skill
to prevent collisions between instruments as well as with the surrounding
tissues and organs. This procedure can be particularly challenging
when considering the reduced workplace, the reduced haptic perception,
and the 2D endoscopic vision in pediatric surgery.

\subsection{Related works}

Many studies have been conducted on the generation of guidance virtual
fixtures for suturing. For instance, Kapoor \emph{et al.} used guidance
virtual fixtures to guide needle insertion \cite{kapoor2005spatial}
and in bimanual knot placement \cite{Kapoor2008}. Chen \emph{et al.}
\cite{chen2016virtual} introduced a knot-tying virtual fixture by
constraining the tooltip to a plane. Fontanelli \emph{et al.} \cite{Fontanelli2018}
compared assistive methods for needle insertion and developed a guidance
virtual fixture to constrain the position of the instrument along
a specific trajectory while the rotations are free. Selvaggio \emph{et
al.} \cite{Selvaggio2019} proposed a haptic-guided shared control
for needle grasping that significantly improved needle re-grasping
performance. The looping task differs from needle insertion in that
both instruments have to interact with each other and, especially
inside the constrained workspace inside the infant, collisions have
to be carefully taken into account.

To facilitate safer robot-assisted minimally-invasive partial nephrectomy,
Banach \emph{et al. }\cite{Banach2019} proposed tool-shaft and anatomy
collision avoidance using the elastoplastic frictional force control
model.

Looi \emph{et al.} \cite{Looi2013} showed a proof of concept of a
robot for image-guided anastomosis in pediatric/neonate surgery. The
authors reported that the robot was able to autonomously perform sutures
in some scenarios, but had difficulties in more realistic situations
owing to the workspace restrictions.

In prior works, our group has focused on the generation of dynamic
regional virtual fixtures to prevent collisions between instruments
and to generate task constraints using vector-field inequalities \cite{Marinho2019}.
More recently, we applied vector-field inequalities to teleoperation
tasks and developed a unified framework for teleoperation \cite{Marinho2019a}.
Those works included simulations and experiments, in which the relevant
task constraints were appropriately maintained. The generation of
guidance virtual fixtures, i.e. specific constraints to optimize task
execution, has not yet been explored using our framework.

\subsection{Statement of contributions}

In this letter, we briefly establish the benefits of the vector field
inequalities (VFI) method over competing frameworks in the context
of real-time virtual-fixtures generation (Section~\ref{subsec:Why-vector-field-inequalities?})
and propose an assistive method based on virtual fixtures to assist
in the looping task in pediatric/neonatal surgery with the following
components (Section~\ref{sec:Proposed-method}). The proposed assistive
method is evaluated in simulations and experiments with naive and
expert users using an anatomically correct infant model. \textcolor{black}{To
the best of the authors' knowledge, this is the first work proposing
guidance virtual fixtures for the looping task.}

\section{Mathematical Background}

In our proposed method, virtual fixtures are modeled using dual quaternion
algebra and the VFI method \cite{Marinho2019} based on quadratic
programming for closed-loop inverse kinematics. In this section, we
briefly present the required mathematical background and notation.
The proposed technique for assistance is shown in Section~\ref{sec:Proposed-method}.

\subsection{Quaternions and dual quaternions\protect\footnote{We use the notation of \cite{Marinho2019}.}}

The quaternion set is $\mathbb{H}\triangleq\left\{ \quat h=h_{1}+\imi h_{2}+\imj h_{3}+\imk h_{4}\,:\,h_{1},h_{2},h_{3},h_{4}\in\mathbb{R}\right\} $,
in which the imaginary units $\imi$, $\imj$, and $\imk$ obey $\hat{\imath}^{2}=\hat{\jmath}^{2}=\hat{k}^{2}=\hat{\imath}\hat{\jmath}\hat{k}=-1$.
Elements of the set $\mathbb{H}_{p}\triangleq\left\{ \quat p\in\mathbb{H}\,:\,\real{\quat p}=0\right\} $
represent points in $\mathbb{R}^{3}$. The operator $\vecthree{\quat p}$
maps a $\quat p\in\mathbb{H}_{p}$ to $\mathbb{R}^{3}$. The set of
quaternions with unit norm, $\mathbb{S}^{3}\triangleq\left\{ \quat{\quat r}\in\mathbb{H}\,:\,\norm{\quat{\quat r}}=1\right\} $,
represent the rotation $\quat r=\cos\left(\phi/2\right)+\quat v\sin\left(\phi/2\right)$,
where $\phi\in\mathbb{R}$ is the rotation angle around the rotation
axis $\quat v\in\mathbb{S}^{3}\cap\mathbb{H}_{p}$. The operator $\vecfour{\quat h}$
maps a $\quat h\in\mathbb{H}$ to $\mathbb{R}^{4}$.

The dual quaternion set is $\mathbb{\mathcal{H}}\triangleq\left\{ \dq h=h+\varepsilon h'\,:\,h,h'\in\mathbb{H},\varepsilon^{2}=0,\varepsilon\ne0\right\} $,
where $\varepsilon$ is the dual unit. Elements of the set $\mathbf{\mathcal{\underline{S}}}\triangleq\left\{ \dq x\in\mathcal{H}\,:\,\norm{\dq x}=1\right\} $
represent poses $\dq x=\quat r+\varepsilon(1/2)\quat t\quat r$, where
$\quat r\in\mathbb{S}^{3}$ is the rotation and $\quat t\in\mathbb{H}_{p}$
is the translation.

Elements of the set $\mathcal{H}_{p}\triangleq\left\{ \dq h_{p}\in\mathcal{H}\,:\,\real{\dq h}=0\right\} $
are called pure dual quaternions and represent points $\left(\mathcal{H}_{p}\supset\mathbb{H}_{p}\right)$,
lines, and planes in $\mathbb{R}^{3}$. 

Given $\dq a,\dq b\in\mathcal{H}_{p}$, their inner product and cross
product are respectively $\dotmul{\dq a}{\dq b}\triangleq-0.5\left(\dq a\dq b+\dq b\dq a\right)$
and $\dq a\times\dq b\triangleq0.5\left(\dq a\dq b-\dq b\dq a\right).$

A Plücker line can be written as $\dq l=\quat l+\varepsilon\quat m$,
where $\quat l\in\mathbb{H}_{p}\cap\mathbb{S}^{3}$ represents the
line direction, and $\quat m=\quat{p_{l}\times\quat l}$ is the line
moment, in which $\quat{p_{l}}\in\mathbb{H_{P}}$ is a point in the
line.

A plane can be written as $\dq{\pi}\triangleq\quat{n_{\pi}}+\varepsilon d_{\pi}$,
where $\quat{n_{\pi}}\in\mathbb{H}_{p}\cap\mathbb{S}^{3}$is the normal
to the plane, and $d_{\pi}=\dotmul{\quat p_{\pi}}{\quat n_{\pi}}\in\mathbb{R}.$

\subsection{Constrained optimization algorithm\label{subsec:unified}}

Without loss of generality, the following constrained optimization
algorithm can be used to teleoperate two identical slave robots $R_{i}$
with $i=1,2$ \cite{Marinho2019a}:
\begin{alignat}{1}
\min_{\dot{\myvec q}}\  & \beta\mathscr{F}_{1}+\left(1-\beta\right)\mathscr{F}_{2}\label{eq:quadratic_problem_teleoperation}\\
\text{subject to}\  & \mymatrix W\dot{\myvec q}\preceq\myvec w,\nonumber 
\end{alignat}
where
\begin{equation}
\mathscr{F}_{i}\triangleq\alpha f_{t,i}+\left(1-\alpha\right)f_{r,i}+f_{\Lambda,i},
\end{equation}
in which $f_{t,i}\triangleq\norm{\mymatrix J_{i,\quat t}\dot{\myvec q}_{i}+\eta\vecthree{\tilde{\myvec t}_{i}}}_{2}^{2}$,
$f_{r,i}\triangleq\norm{\mymatrix J_{i,\quat r}\dot{\myvec q}_{i}+\eta\vecfour{\tilde{\myvec r}_{i}}}_{2}^{2}$,
and $f_{\Lambda,i}\triangleq\norm{\mymatrix{\Lambda}\dot{\myvec q}_{i}}_{2}^{2}$
are the unweighted cost functions related to the end-effector translation,
end-effector rotation, and joint velocities of the $i$-th robot,
respectively. Furthermore, each $i$-th robot has a vector of joint
velocities $\dot{\myvec q}_{i}$ $\in\mathbb{R}^{n_{i}}$, a translation
Jacobian $\mymatrix J_{i,\quat t}$ $\in\mathbb{R}^{3\times n_{i}}$,
a translation error $\tilde{\quat t}_{i}\triangleq\quat t_{i}-\quat t_{i,d}$
$\in$ $\mathbb{H}_{p}$, a rotation Jacobian $\mymatrix J_{i,\quat r}$
$\in\mathbb{R}^{4\times n_{i}}$, and a switching rotational error
$\tilde{\quat r}_{i}$ $\in$ $\mathbb{H}_{p}$. In addition, $\dot{\myvec q}=\begin{bmatrix}\dot{\myvec q}_{1}^{T} & \dot{\myvec q}_{2}^{T}\end{bmatrix}^{T}$,
and $\mymatrix{\Lambda}\in\mathbb{R}^{\left(n_{1}+n_{2}\right)\times\left(n_{1}+n_{2}\right)}$
is a positive definite damping matrix. Finally, $\alpha,\beta$ $\in$
$\left[0,1\right]$ are weights used to define the priorities between
the translation and the rotation and the priorities between robots.
The linear constraints $\mymatrix W\dot{\myvec q}\preceq\myvec w$
can be used to avoid joint limits \cite{cheng1994} and to generate
task constraints \cite{Marinho2019}. Each parameter is explained
in more detail in \cite{Marinho2019a}.

\subsection{VFI method}

The VFI method \cite{Marinho2019} uses a function $d\triangleq d(\myvec q,t)$
$\in$ $\mathbb{R}$ that represents the (signed) distance between
two geometric primitives. The time-derivative of the distance is 
\begin{equation}
\dot{d}=\underbrace{\partialfrac{d(\myvec q,t)}{\myvec q}}_{\mymatrix J_{d}}\dot{\myvec q}+\zeta(t),\label{eq:dynamic-vector-field-inequalities}
\end{equation}
where $\mymatrix J_{d}$ $\in$ $\mathbb{R}^{n}$ is the distance
Jacobian and $\zeta(t)=\dot{d}-\boldsymbol{J}_{d}\dot{\quat q}$ is
the residual. \textcolor{black}{Moreover, let there be a safe distance
$d_{\text{safe}}\triangleq d_{\text{safe}}(t)$ $\in$ $\left[0,\infty\right)$
and an error $\tilde{d}\triangleq\tilde{d}(\myvec q,t)=d-d_{\text{safe}}$
to generate restricted zones or $\tilde{d}\triangleq d_{\text{safe}}-d$
to generate safe zones.}

\textcolor{black}{With these definitions, and given $\eta_{d}$ $\in$
$\left[0,\infty\right)$, the signed distance dynamics is constrained
by $\dot{\tilde{d}}\geq-\eta_{d}\tilde{d}$, which actively filters
the robot motion only in the direction approaching the restricted
zone boundary so that the primitives do not collide. }

The following constraint is used to generate restricted zones
\begin{equation}
-\mymatrix J_{d}\dot{\myvec q}\leq\eta_{d}\tilde{d}+\zeta_{\text{safe}}\left(t\right),
\end{equation}
for $\zeta_{\text{safe}}\left(t\right)\triangleq\zeta\left(t\right)-\dot{d}_{\text{safe}}$.
Alternatively, safe zones are generated by
\begin{equation}
\mymatrix J_{d}\dot{\myvec q}\leq\eta_{d}\tilde{d}-\zeta_{\text{safe}}\left(t\right).
\end{equation}

\subsubsection{Generating an entry sphere using VFIs\label{subsec:Generating-an-entry-sphere}}

As an example relevant to the application of this letter, in infant
surgery, instead of an entry-point constraint, an entry-sphere constraint
is used \cite{Marinho2019a}. This constraint replicates the manual
technique of medical doctors that utilizes the compliance of the infant's
skin to increase the reachable workspace. To generate this constraint,
without loss of generality, let $\dq x_{i}=\quat r_{i}+\varepsilon\frac{1}{2}\quat t_{i}\quat r_{i}$
be a coordinate frame whose $z-$axis is along the shaft of the instrument
of a given robot $R_{i}$. The Plücker line associated with the instrument
shaft's axis, $\dq l_{z,i}\mathcal{\underline{S}},$ can be expressed
by
\begin{equation}
\dq l_{z,i}=\quat l_{z,i}+\varepsilon\quat m_{z,i}
\end{equation}
in which $\quat l_{z,i}=\quat r_{i}\hat{k}\conj{\quat r_{i}}\in\mathbb{H}_{p}\cap\mathbb{S}^{3}$
is the line direction and $\quat m_{z,i}=\quat t_{i}\times\quat l_{z,i}\in\mathbb{H}_{p}$
is the line moment. With the center of the entry sphere given by $\quat p_{\text{rcm},i}\in\mathbb{\mathbb{H}}_{p}$,
the derivative of the squared distance between the entry point and
the instrument's shaft is given by
\begin{equation}
\dot{D}_{\text{rcm},i}=\distancejacobian{l_{z,i}}p{}\myvec{\dot{q}}_{i},
\end{equation}
where $\distancejacobian{l_{z,i}}p{}\in\mathbb{R}^{n_{i}}$ \textcolor{black}{is
the line-to-point squared distance Jacobian }\cite{Marinho2019}\textcolor{black}{.
Using the VFI method, the instrument's shaft can be constrained by
an entry sphere of squared radius $D_{\text{safe,rcm,i}}\in\mathbb{R}$
by using the following linear constraint:}
\begin{equation}
\underbrace{\distancejacobian{l_{z,i}}p{}\myvec{\dot{q}}_{i}}_{\mymatrix W_{\text{rcm}}}\leq\underbrace{\eta_{\text{rcm},i}(D_{\text{safe,rcm},i}-D_{\text{rcm},i})}_{\myvec w_{\text{rcm}}},\label{eq:RCM_inequality}
\end{equation}
where $\eta_{rcm,i}\in[0,\infty)$ is a gain that affects the allowed
speed of the instrument's shaft toward the surface of the sphere.

\section{Why VFI?\label{subsec:Why-vector-field-inequalities?}}

\begin{figure}[h]
\begin{centering}
\includegraphics[width=0.9\columnwidth]{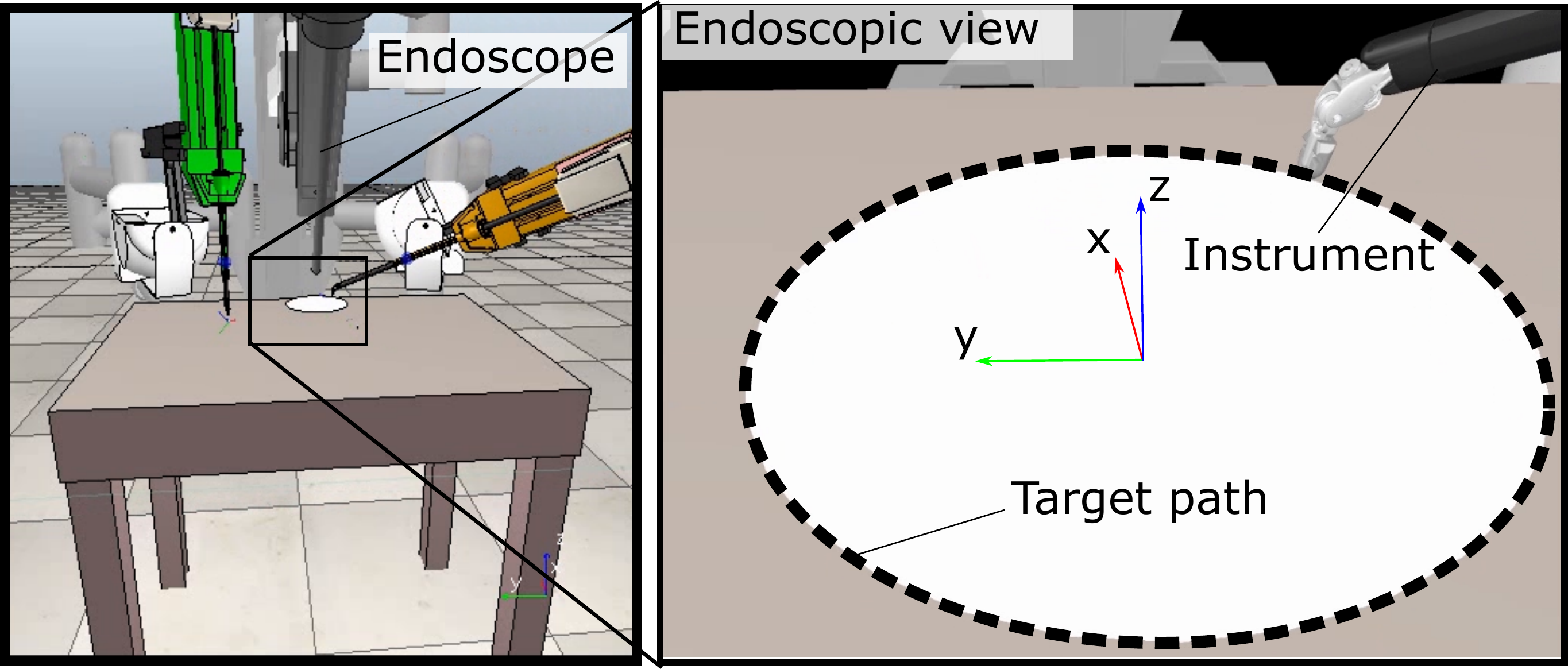}
\par\end{centering}
\caption{\label{fig:simulationsetup} The V-REP simulation used in the simulation
comparison. The circle's $z$-axis position changed in time following
a sinusoidal wave $d\left(t\right)=d_{o}+0.01\sin\left(2\pi0.1t\right)$.}
\end{figure}

There are myriad of competing techniques for the generation of virtual
fixtures/active constraints \cite{bowyer2014active}. For nonredundant
robots, such as the da Vinci, virtual fixtures based on force feedback
on the master side are effective \cite{bowyer2013dynamic,bowyer2014active,Banach2019}.
For redundant robots, such as the SmartArm and similar systems, only
force feedback on the master side is in general not enough, owing
to possibly infinite mappings between master and slave postures. That
is, with unconstrained inverse kinematics, pushing the master in a
given direction, in general, \emph{does not guarantee} that a redundant
slave's links will move in a repeatable manner. This problem becomes
more evident in surgeries in constrained spaces such as pediatric
and neonate surgery. In this context, we have been developing a framework
\cite{Marinho2019a} based on active constraints generated through
constrained optimization on the slave side, which \emph{guarantees}
the integrity of the robotic systems and the safety of the patients
and operating room personnel. On the master side, we add Cartesian
impedance to make the operator aware of the workspace constraints.

To validate the benefits of the VFI method over existing works, we
show a brief simulation study.

\textcolor{black}{In the existing literature, active constraints using
constrained optimization in the context of robotic surgery were initially
proposed by Kapoor }\textcolor{black}{\emph{et al}}\textcolor{black}{.
\cite{kapoor2006constrained}. They proposed several primitives that
can be used to assemble customizable virtual fixtures, and one of
their primitives in common with the VFI method is the plane constraint.
In this context, we compare the performance of the constraint proposed
in \cite{kapoor2006constrained} with what can be achieved with VFIs.
Because there is no explicit objective function for teleoperation
in \cite{kapoor2006constrained}, we used the same objective function
(with the same gains) for both methods and changed only the constraints.}

\textcolor{black}{To compare both techniques, we used the V-REP simulation
developed in \cite{Fontanelli2018}. A suitable scenario using the
plane constraint common to both techniques is to require the robots'
tooltips to be restricted to a dynamic plane while the robot is teleoperated.
The dynamic-plane distance changes in a sinusoidal manner along the
$z$-axis according to $d\left(t\right)=d_{o}+0.01\sin\left(2\pi0.1t\right)$
with a fixed normal.}

\textcolor{black}{The user was asked to trace, by using the master
interface, a circle seen through the simulated endoscope. The experiment
was performed once without any plane motion, and that trajectory was
recorded and played back for each technique to ensure that the trajectory
on the $xy$-plane was the same for all cases. To further increase
the difficulty of the task, the tooltip starts $30$~mm away from
the dynamic plane.}

\begin{figure}[h]
\begin{centering}
\includegraphics[width=0.9\columnwidth]{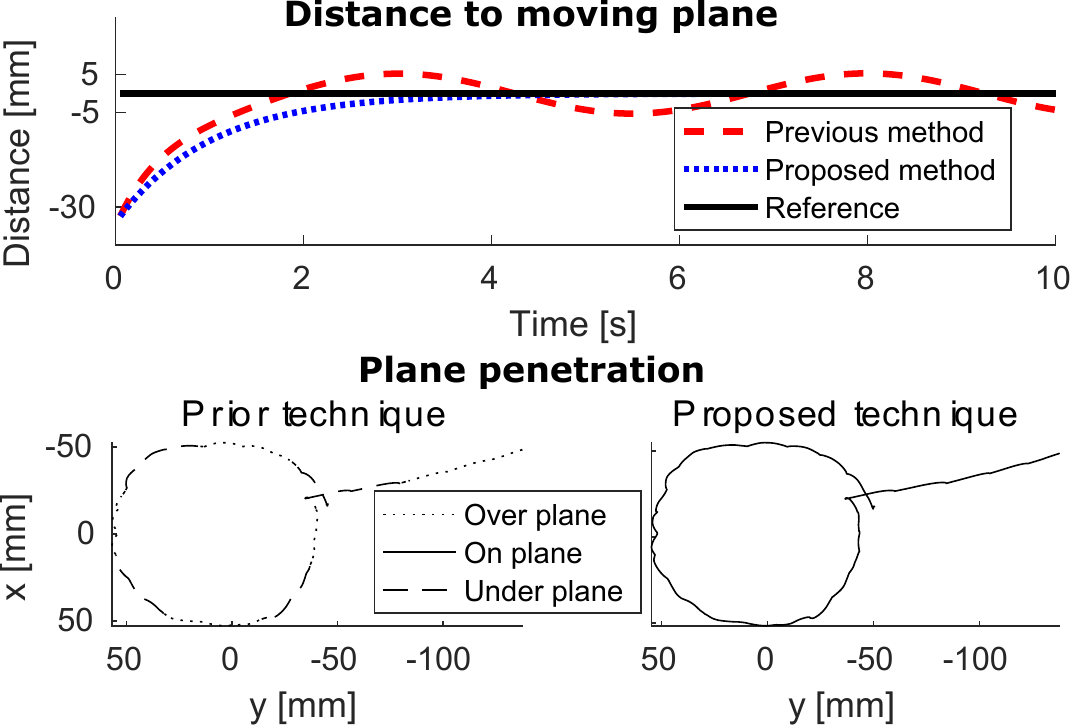}
\par\end{centering}
\caption{\label{fig:planedistance}Robot tooltip's distance to the moving plane.
When the plane constraint is defined by using the prior technique
\cite{kapoor2006constrained}, the tooltip considerable deviates from
the plane. Using our proposed technique, the tooltip converges to
the moving plane and moves with it.}
\end{figure}

\textcolor{black}{The tooltip distance with respect to the plane for
each method is shown in Fig.~\ref{fig:planedistance}. A considerable
deviation from the plane, $5.27$~mm, was observed when using the
constraints proposed in \cite{kapoor2006constrained}. This occurred
because the constraint proposed in \cite{kapoor2006constrained} does
not take into account the instantaneous velocity of the plane; therefore,
there was a steady-state offset between the desired plane and the
actual plane. When VFIs were used, the residual (as shown in \eqref{eq:dynamic-vector-field-inequalities})
acted as a feed-forward term that compensated for the plane's instantaneous
velocity, which allowed convergence to the moving plane.}

This property is required for the proper generation of dynamic virtual
fixtures such as the ones proposed in this letter.

\section{Proposed assistive method for suturing\label{sec:Proposed-method}}

The proposed assistive method for suturing is divided into two parts:
a constrained optimization algorithm on the slave side (Section~\ref{subsec:constrained_optimization})
and Cartesian impedance feedback on the master side (Section~\ref{subsec:cartesian-impedance}).

\subsection{Problem statement}

\begin{figure}[h]
\centering

\noindent\resizebox{0.9\columnwidth}{!}{%

\huge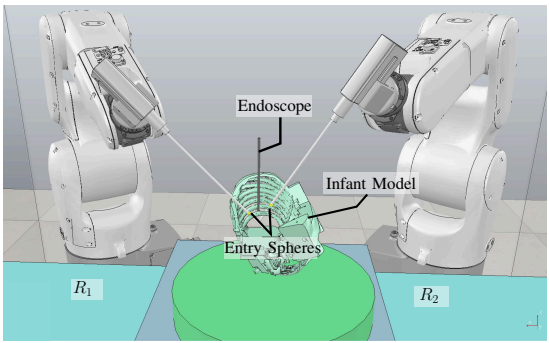

}

\caption{\label{fig: vrep-setup} Robot setup.}
\end{figure}

Consider the setup shown in Fig.~\ref{fig: vrep-setup}. Let there
be two robots, $R_{1}$ and $R_{2}$, with instruments as their end
effector. Suppose that the instruments can be simplified as capsules
(cylinders with spheres on their endpoints). For robot-aided endoscopic
infant surgery, each instrument has to be inserted through the intercostal
space (between the ribs) of the infant. Each incision on the skin
cannot be pushed, to prevent further damage to the tissue. This is
accomplished by adding an entry-sphere constraint for each robot,
as discussed in Section~\ref{subsec:Generating-an-entry-sphere}.

One of the required steps in suturing is to loop the thread about
one of the instruments before grasping the loose end of the thread
and tying the knot. In this step, the inexperienced user can loop
too far (risking collisions with the anatomy) or too close (risking
collisions with the other instrument). The proposed technique, described
in the following sections, aims to assist the surgeon in performing
the looping in a safer and more controlled manner.

\subsection{Slave side: Constrained optimization\label{subsec:constrained_optimization}}

Let $R_{1}$ be the slave robot operated by the non-dominant hand,
and $R_{2}$ be the robot operated by the dominant hand. In this work,
we propose a set of dynamic virtual fixtures attached to $R_{1}$
that constrain the motion of $R_{2}$ to aid the robotic thread looping
task in suturing. The proposed virtual fixtures have been designed
by careful inspection of videos of medical practice in pediatric surgery
\cite{Harada2016} and fruitful discussions with cooperating surgeons.
We make two basic assumptions. First, restricting the motion of one
instrument with respect to the other using virtual fixtures during
the looping task to a guidance region can be helpful in reducing extraneous
motion. Second, adding a guidance surface can further improve task
performance.

\begin{figure}[h]
\centering

\def\svgwidth{0.8\columnwidth}

\scriptsize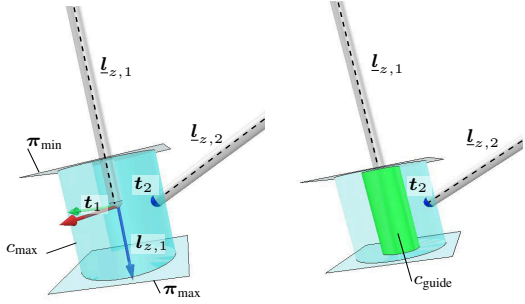

\caption{\label{fig:GVF_TG} Visualization of LVFs (left) and TGC (right).}
\end{figure}

The dynamic virtual fixtures, which we call looping virtual fixtures
(LVFs), are generated by employing a shaft\textemdash shaft collision
avoidance primitive \cite{Marinho2019} plus three geometric primitives
for the tooltip of $R_{1}$. First, we attach a dynamic cylinder $c_{\text{max}}$with
a radius of $r_{\text{max}}$ around the $z-$axis of the end effector
of $R_{1}$
\begin{equation}
\dq l_{z,1}=\overbrace{\quat r_{1}\hat{k}\conj{\quat r_{1}}}^{\quat l_{z,1}}+\dual\left(\quat t_{1}\times\quat l_{z,1}\right)
\end{equation}
in which $\quat r_{1}\in\mathbb{S}^{3}$ and $\quat t_{1}\in\mathbb{H}_{p}$
are respectively the rotation and translation of $R_{1}$. The cylinder
is cut with a pair of planes $\pi_{\text{min}}$ and $\pi_{\text{max}}$whose
normals are collinear with $\dq l_{z,1}$ and are respectively placed
at $d_{\pi,\text{min}}$ and $d_{\pi,\text{max}}$ from the tooltip
of $R_{1}$
\begin{align}
\pi_{\min}\triangleq & \quat n_{\pi}+\varepsilon\left(\dotmul{\quat t_{1}}{\quat n_{\pi}}+d_{\pi,\text{min}}\right)\\
\pi_{\max}\triangleq & \quat n_{\pi}+\varepsilon\left(\dotmul{\quat t_{1}}{\quat n_{\pi}}+d_{\pi,\max}\right)
\end{align}
in which $\quat n_{\pi}=\quat r_{1}\hat{k}\conj{\quat r_{1}}$. These
geometric constraints, shown in Fig.~\ref{fig:GVF_TG}, limit the
motion of $R_{2}$ so that its tooltip is constrained within a motion
envelope to prevent large motions that can be detrimental to task
performance, as well as preventing collisions between the shafts and
the surrounding tissues. The radius of the cylinder and the placement
of the planes can be tuned to balance loop size and task performance.

\textcolor{black}{The LVF can be generated using linear constraints
by employing VFIs. With the shaft-shaft collision avoidance given
by \cite[Eq. 64]{Marinho2019}
\begin{equation}
\mymatrix W_{\text{ss}}\leq\myvec w_{\text{ss}},
\end{equation}
the LVFs can be generated by the following linear constraints
\begin{gather}
\underbrace{\left[\begin{array}{cc}
\mymatrix W_{\text{ss}}\\
\distancejacobian{l_{z,1}}{t_{2}}{} & \distancejacobian{t_{2}}{l_{z,1}}{}\\
-\distancejacobian{\pi_{\min}}{t_{2}}{} & -\distancejacobian{t_{2}}{\pi_{\min}}{}\\
\distancejacobian{\pi_{\max}}{t_{2}}{} & \distancejacobian{t_{2}}{\pi_{\max}}{}
\end{array}\right]}_{\mymatrix W_{\text{LVF}}}\leq\eta_{d}\underbrace{\left[\begin{array}{c}
\myvec w_{\text{ss}}\\
(r_{\max})^{2}-\squaredistance{l_{z,1}}{t_{2}}{}\\
\distance{\pi_{\text{min}}}{t_{2}}{}-(d_{\pi,\text{min}})\\
(d_{\pi,\text{max}})-\distance{\pi_{\text{max}}}{t_{2}}{}
\end{array}\right]}_{\myvec w_{\text{LVF}}},\label{eq:LVF_constraints}
\end{gather}
in which $\distancejacobian{l_{z,1}}{t_{2}}{}$ is the line-to-point
distance Jacobian \cite[Eq. 34]{Marinho2019} between the Plücker
line collinear with the shaft of $R_{1}$, $\dq l_{z,1}$, and the
tooltip of $R_{2}$, $\quat t_{2}$. Moreover, $\distancejacobian{t_{2}}{l_{z,1}}{}$
is the point-to-line distance Jacobian \cite[Eq. 32]{Marinho2019}
between $\dq l_{z,1}$ and $\quat t_{2}$. Furthermore, $\distancejacobian{\pi_{\min}}{t_{2}}{}$
and $\distancejacobian{\pi_{\max}}{t_{2}}{}$ are plane-to-point distance
Jacobians \cite[Eq. 56]{Marinho2019} between $\pi_{\text{min}}$
and $\quat t_{2}$, and $\pi_{\text{max}}$ and $\quat t_{2}$, respectively.
Conversely, $\distancejacobian{t_{2}}{\pi_{\min}}{}$ and $\distancejacobian{t_{2}}{\pi_{\max}}{}$
are the point-to-plane distance Jacobians \cite[Eq. 59]{Marinho2019}
between $\pi_{\text{min}}$ and $\quat t_{2}$, $\pi_{\text{max}}$
and $\quat t_{2}$, respectively. $\squaredistance{l_{z,1}}{t_{2}}{}$
is the line-to-point squared distance \cite[Eq. 33]{Marinho2019}
between $\dq l_{z,1}$ and $\quat t_{2}$ . Lastly, $\distance{\pi_{\text{min}}}{t_{2}}{}$
and $\distance{\pi_{\text{max}}}{t_{2}}{}$ are the plane-to-point
distances \cite[Eq. 54]{Marinho2019} between $\pi_{\text{min}}$
and $\quat t_{2}$, and $\pi_{\text{max}}$ and $\quat t_{2}$, respectively.}

To further increase assistance, we propose a guidance virtual fixture,
called trajectory guidance cylinder (TGC). It comprises a cylinder
$c_{\text{guide}}$ with its centerline collinear to $\dq l_{z,1}$
and radius $r_{\text{guide }}$that is placed inside the LVFs and
used to guide the tooltip of $R_{2}$.

We propose the following constrained optimization problem to implement
both the LVFs and the TGC
\begin{gather}
\min_{\dot{\myvec q}}\ \gamma(\mathscr{F}_{1}+\mathscr{F}_{2})+(1-\gamma)(\mathscr{G}_{1}+\mathscr{G}_{2})+(f_{\Lambda,1}+f_{\Lambda,2})\nonumber \\
\text{subject to}\ \mymatrix W\dot{\myvec q}\preceq\myvec w,\label{eq:proposed_optimization}
\end{gather}
where $\mathscr{F}_{i}$ is related to trajectory tracking as in \eqref{eq:quadratic_problem_teleoperation}
and $\mathscr{G}_{i}$ is related to the TGC of the $i$-th robot
as follows
\begin{align}
\mathscr{G}_{1}\triangleq & \distancejacobian{l_{z,1}}{t_{2}}{}\dot{\myvec q}_{1}+\eta_{\text{guide}}\tilde{D}_{\text{guide}}\\
\mathscr{G}_{2}\triangleq & \distancejacobian{t_{2}}{l_{z,1}}{}\dot{\myvec q}_{2}+\eta_{\text{guide}}\tilde{D}_{\text{guide}},
\end{align}
where the guidance error is defined as $\tilde{D}_{\text{guide}}\triangleq\squaredistance{l_{z,1}}{t_{2}}{}-r_{\text{guide}}^{2}$.
Lastly, $\alpha,\gamma$ $\in$ $\left[0,1\right]$ are, respectively,
weights used to define the priorities between the translation and
the rotation and between the master-slave tracking and the TGC. The
linear constraints are given by
\begin{gather}
\overbrace{\left[\begin{array}{c}
\mymatrix W_{\text{JL}}\\
\mymatrix W_{\text{rcm}}\\
\mymatrix W_{\text{LVF}}
\end{array}\right]}^{\mymatrix W}\leq\overbrace{\left[\begin{array}{c}
\myvec w_{\text{JL}}\\
\eta_{\text{rcm}}\myvec w_{\text{rcm}}\\
\eta_{d}\myvec w_{\text{LVF}}
\end{array}\right]}^{\myvec w},
\end{gather}
in which $\mymatrix W_{\text{JL}}\leq\myvec w_{\text{JL}}$ is related
to joint limit avoidance \cite{cheng1994}, $\mymatrix W_{\text{rcm}}\leq\myvec w_{\text{rcm}}$
is related to the entry-sphere constraint as in \eqref{eq:RCM_inequality},
and finally $\mymatrix W_{\text{LVF}}\leq\myvec w_{\text{LVF}}$ as
defined in \eqref{eq:LVF_constraints}.

\subsection{Master side: Cartesian impedance with guidance\label{subsec:cartesian-impedance}}

\begin{figure}[h]
\centering

\def\svgwidth{0.8\columnwidth}

\scriptsize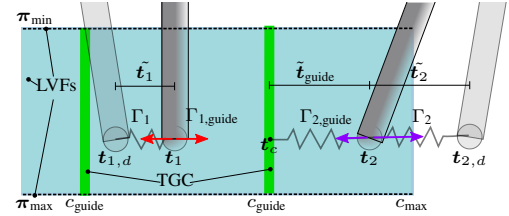

\caption{\textcolor{black}{\label{fig:FB} Visualization of the Cartesian impedance
from the point-of-view of the slave robots. The blue region is the
safe region enveloped by the LVFs and the green line represents the
TGC. The forces $\Gamma_{1},$$\Gamma_{2}$, $\Gamma_{1,\text{guide}}$,
and $\Gamma_{2,\text{guide}}$ are calculated based on the slave-side
errors, projected to the master side, and applied at the master interfaces
according to \eqref{eq:master_1_ff} and \eqref{eq:master_2_ff}.
The guidance error is $\tilde{\protect\myvec t}_{\text{guide}}\triangleq\protect\quat t_{2}-\protect\quat t_{c}$
in which $\protect\quat t_{c}$ is the translation of the point in
$c_{\text{guide}}$ closest to $\protect\quat t_{2}$.}}
\end{figure}

\textcolor{black}{In addition to the existing method of Cartesian
force feedback introduced in \cite{Marinho2019a}, we propose additional
force feedback to guide the tooltip of $R_{2}$ to the cylinder $c_{\text{guide}}$
in the form
\begin{align}
\quat{\Gamma}_{1,\text{total}}^{\text{master}} & \triangleq\overbrace{-\eta_{f}\gamma\tilde{\myvec t}_{1}^{\text{master}}}^{\Gamma_{1}^{\text{master}}}\overbrace{-\eta_{f}\left(1-\gamma\right)\tilde{\myvec t}_{\text{guide}}^{\text{master}}}^{\Gamma_{1,\text{guide}}^{\text{master}}}-\eta_{V}\dot{\myvec t}_{1,\text{master}}^{\text{master}},\label{eq:master_1_ff}\\
\quat{\Gamma}_{2,\text{total}}^{\text{master}} & \triangleq\underbrace{-\eta_{f}\gamma\tilde{\myvec t}_{2}^{\text{master}}}_{\Gamma_{2}^{\text{master}}}\underbrace{+\eta_{f}\left(1-\gamma\right)\tilde{\myvec t}_{\text{guide}}^{\text{master}}}_{\Gamma_{2,\text{guide}}^{\text{master}}}-\eta_{V}\dot{\myvec t}_{2,\text{master}}^{\text{master}}.\label{eq:master_2_ff}
\end{align}
For each $i-$th robot master\textendash slave pair, $\quat{\Gamma}_{i,\text{total}}^{\text{master}}$
is the force on the $i-$th master and $\eta_{f},\eta_{V}$ $\in$
$\left(0,\infty\right)$ are stiffness and viscosity parameters. $\tilde{\myvec t}_{i}^{\text{master}}$
is the translation error of the $i-$th slave seen from the point
of view of the master,}\footnote{\textcolor{black}{The point-of-view of the master is the point-of-view
of the endoscopic camera.}}\textcolor{black}{{} and $\dot{\myvec t}_{i,\text{master}}^{\text{master}}$
is the linear velocity of the $i-$th master interface. $\tilde{\myvec t}_{\text{guide}}^{\text{master}}$
is the guidance error of the slaves from the point of view of the
master. Finally, $\gamma$ $\in$ $\left[0,1\right]$ is used to define
the priority between the master-slave tracking force feedback, $\Gamma_{i}^{\text{master}}$,
and the TGC force feedback, $\Gamma_{i,\text{guide}}^{\text{master}}$.
Those two forces act as a spring, trying to reduce the tracking and
guidance errors, as shown in Fig.~\ref{fig:FB}.}

\section{Simulation and Experiment\label{sec:Experiments}}

To evaluate the technique proposed in this paper, we designed a simulation
study and an experimental study. The simulation entailed naive users
operating a simulator (V-REP, Coppelia Robotics, Switzerland) under
three different conditions to evaluate the effects of the proposed
technique. The experimental study investigated medical doctors' performance
using the proposed technique while operating a robotic system \cite{Kim2018}
based on the SmartArm architecture \cite{marinho2019integration},
with a two degrees-of-freedom (rotation, grasping) instrument attached,
in a setup equivalent to \cite{Marinho2019a}.

\begin{figure}[h]
\centering

\def\svgwidth{1.0\columnwidth}

\scriptsize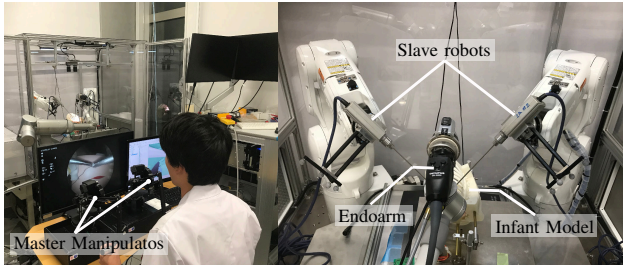

\caption{\label{fig: experiment-setup} Master\textendash slave robotic system.
Master-side (\emph{left}), slave-side (\emph{right}).}
\end{figure}

\begin{figure}[h]
\centering

\def\svgwidth{1.0\columnwidth}

\scriptsize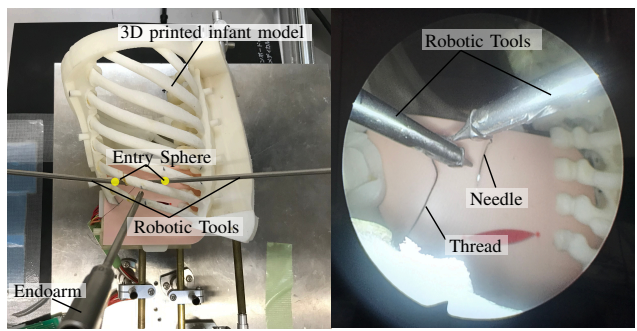
\centering{}\caption{\label{fig:nfant-model-1} Experimental Setup. Top view of the 3D
printed infant model \cite{Takazawa2018} (\emph{left}) and endoscopic
view (\emph{right}).}
\end{figure}

The experimental setup is shown in Fig.~\ref{fig: experiment-setup}.
The 3D model of an infant \cite{Harada2016} (Fig.~\ref{fig:nfant-model-1})
was placed between the two robotic arms, and the entry spheres (Section~\ref{subsec:Generating-an-entry-sphere})
were placed between the ribs of the infant model at their relevant
locations. In these experiments, the two robotic arms (DENSO VS050,
DENSO WAVE Inc., Japan) were equipped with rigid 3.5-mm-diameter instruments.
The simulator replicated the experimental setup.

Both the simulation and the experiment used the same software implementation
described in \cite{marinho2019integration}.

All p values reported in this section were obtained through the (two-tailed)
Wilcoxon signed-rank task.

\begin{table}[h]
\caption{\label{tab:all-parameters} Control parameters and virtual fixture
parameters.}

\centering

\noindent\resizebox{1.0\columnwidth}{!}{%
\begin{centering}
\textcolor{black}{}%
\begin{tabular}{ccccccccccc}
\hline 
\noalign{\vskip\doublerulesep}
$\alpha$ & $\beta$ & \textcolor{black}{$\gamma$} & \textcolor{black}{$\eta$} & \textcolor{black}{$\eta_{\text{d}}$} & \textcolor{black}{$\eta_{\text{rcm}}$} & $\eta_{\text{guide}}$ & \textcolor{black}{$\Lambda$} & \textcolor{black}{$\eta_{f}$} & \textcolor{black}{$\eta_{V}$} & \textcolor{black}{MS}\tabularnewline[\doublerulesep]
\hline 
\noalign{\vskip\doublerulesep}
\noalign{\vskip\doublerulesep}
\textcolor{black}{0.999} & 0.6 & 0.01 & \textcolor{black}{150} & \textcolor{black}{30} & 30 & 1 & \textcolor{black}{0.02} & 50 & 0.5 & \textcolor{black}{$\unitfrac{1}{2}$}\tabularnewline[\doublerulesep]
\noalign{\vskip\doublerulesep}
\hline 
\noalign{\vskip\doublerulesep}
$r_{\min}$ & $r_{\max}$ & $r_{\text{guide}}$ & $d_{\text{safe,rcm}}$ & $d_{\pi,\min}$ & $d_{\pi,\max}$ &  &  &  &  & \tabularnewline[\doublerulesep]
\hline 
\noalign{\vskip\doublerulesep}
\noalign{\vskip\doublerulesep}
3.5 & 20 & 10 & 2.5 & -8 & 10 &  &  &  &  & \tabularnewline[\doublerulesep]
\noalign{\vskip\doublerulesep}
\end{tabular}
\par\end{centering}
}
\begin{raggedright}
\textcolor{black}{\scriptsize{}$\alpha$: translation error to orientation
error weight (Section~\ref{subsec:unified}).}{\scriptsize\par}
\par\end{raggedright}
\begin{raggedright}
\textcolor{black}{\scriptsize{}$\beta$: robot priority weight (Section~\ref{subsec:unified})}{\scriptsize\par}
\par\end{raggedright}
\begin{raggedright}
\textcolor{black}{\scriptsize{}$\gamma$: }{\scriptsize{}weights between
the master-slave tracking and TGC}\textcolor{black}{\scriptsize{}
(Section~\ref{subsec:constrained_optimization}).}{\scriptsize\par}
\par\end{raggedright}
\begin{raggedright}
\textcolor{black}{\scriptsize{}$\eta$, $\eta_{d}$,$\eta_{\text{guide}}$:
proportional gain of the kinematic controller, LVF, and TGC, respectively.}{\scriptsize\par}
\par\end{raggedright}
\begin{raggedright}
\textcolor{black}{\scriptsize{}$\Lambda$: Robot joint gains (Section~\ref{subsec:unified}).}{\scriptsize\par}
\par\end{raggedright}
\begin{raggedright}
\textcolor{black}{\scriptsize{}$\eta_{F},\eta_{V}$: Cartesian impedance
proportional and viscosity gains, respectively (Section~\ref{subsec:cartesian-impedance}).}{\scriptsize\par}
\par\end{raggedright}
\begin{raggedright}
\textcolor{black}{\scriptsize{}MS: Motion scaling. A motion scaling
of X means that a relative translation of the master was multiplied
by X before being sent to the slave.}{\scriptsize\par}
\par\end{raggedright}
\raggedright{}{\scriptsize{}$r_{\min},r_{\max},r_{\text{guide}},d_{\text{safe,rcm}},d_{\pi,\min},d_{\pi,\max}$
virtual fixture parameters in millimeters (Section}\textcolor{black}{\scriptsize{}~\ref{subsec:constrained_optimization}}{\scriptsize{}).}{\scriptsize\par}
\end{table}

\subsection{\textcolor{black}{Simulation\label{subsec:Simulation}}}

\textcolor{black}{For the simulation study, six volunteers were recruited
among the engineering students at the University of Tokyo, who had
no medical experience. After being shown a video demonstrating the
ideal double loop trajectory, the users were instructed to replicate
that loop as closely as possible. Each user was asked to perform in
the simulator a double loop in a total of five trials in each of the
following three conditions: A1 with only the entry-sphere constraint
(control group), A2 with the entry-sphere constraint and the shaft\textendash shaft
collision avoidance (first mentioned in \cite{Marinho2019}), and
A3 with the entry-sphere constraint and the proposed LVFs and TGC. }

\textcolor{black}{The trials were done using a replicated balanced
Latin-squares design \cite{williams1949experimental} with the trials
and users as blocking factors. Each user performed a total of 15 trials.
To reduce possible biases, the users did not know which condition
was activated at a given trial. The proposed virtual fixtures were
implemented using \eqref{eq:proposed_optimization} and were determined
by pilot studies overseen by a medical doctor. Their relevant parameters
are shown in Table~\ref{tab:all-parameters}.}

\textcolor{black}{The simulation had two purposes. First, to evaluate
the effectiveness of the proposed technique with inexperienced users.
Second, to investigate condition A1, which has no virtual fixtures
to prevent collisions between shafts; therefore, we could investigate
collisions between instruments and the anatomical model without damaging
the physical equipment. It is important to note that the surgical
thread was not simulated; only the loop motion was evaluated.}

\subsubsection{\textcolor{black}{Results and discussion}}

\paragraph{\textcolor{black}{Controller performance}}

\textcolor{black}{To evaluate the performance of the looping task,
we used the task completion time and the error between the tooltip
position of $R_{2}$ and the surface of the guidance cylinder (Section~\ref{subsec:constrained_optimization})
that delineates the ideal looping surface. Table~\ref{tab:all-time-RMSE}
shows the median time and mean error recorded for the six volunteers
for each of the three experimental cases.}

\begin{table}[h]
\caption{\label{tab:all-time-RMSE}User performance in the simulation.}

\centering{}\textcolor{black}{}%
\begin{tabular}{cccc}
\hline 
\noalign{\vskip\doublerulesep}
 & {\small{}A1} & {\small{}A2} & {\small{}A3}\tabularnewline[\doublerulesep]
\hline 
\noalign{\vskip\doublerulesep}
\noalign{\vskip\doublerulesep}
{\small{}Median time {[}s{]}} & {\small{}14.56} & {\small{}17.12} & {\small{}17.10}\tabularnewline[\doublerulesep]
\noalign{\vskip\doublerulesep}
\noalign{\vskip\doublerulesep}
{\small{}Mean error {[}mm{]}} & {\small{}2.267} & {\small{}1.261} & {\small{}1.211}\tabularnewline[\doublerulesep]
\noalign{\vskip\doublerulesep}
\end{tabular}
\end{table}

\textcolor{black}{For A1, the users took a median of 14.56~s to complete
the task. That was the shortest completion time between all experimental
conditions (p < 0.05). However, the mean trajectory error was the
highest and, as expected, there were collisions between the instruments.}

\textcolor{black}{A2 provided a collision-free path for the instruments
and a 44\% reduction in the mean trajectory error with respect to
A1. There was a 18\% increase (p < 0.05) in the median required time
with respect to A1.}

\textcolor{black}{Finally, after adding the proposed guidance virtual
fixtures, in condition A3 the instrument path was also collision-free.
Moreover, there was a reduction of 4\% in the mean trajectory error
with respect to A2 (47\% with respect to A1). There was no reduction
in the median time with respect to A2 (17\% increase with respect
to A1 (p < 0.05)).}

\textcolor{black}{These results show that our proposed method, A3,
is equivalent to A2 in terms of controller performance. It is slightly
superior in terms of mean trajectory error.}

\begin{figure}[tbh]
\centering

\def\svgwidth{0.9\columnwidth}

\footnotesize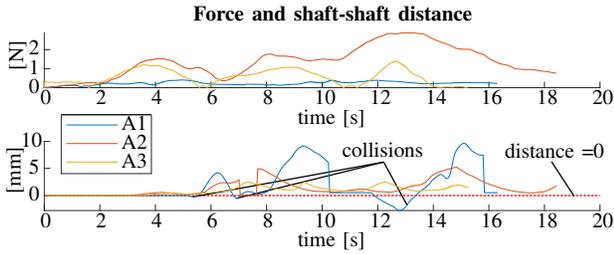

\caption{\label{fig:force} The norm of the force exerted on the right master
interface (\emph{top}) and the signed distance between shafts (\emph{bottom})
for one double-looping trial under each condition on the simulation
study. Only condition A1 had collisions between shafts.}
\end{figure}

\textcolor{black}{A sample of the force rendered on the right master
for each condition is shown in Fig.~\ref{fig:force}.}

\paragraph{\textcolor{black}{User evaluation}}

\begin{figure}[tbh]
\centering

\def\svgwidth{1.0\columnwidth}

\footnotesize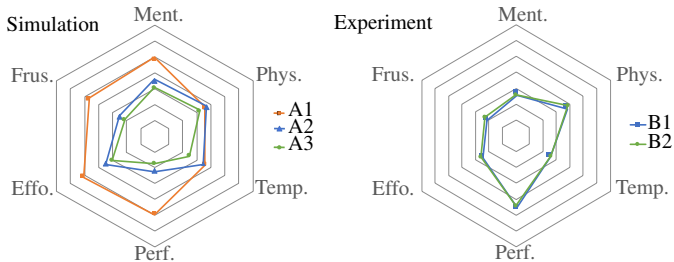

\caption{\label{fig:ex1_NASA TLX}NASA TLX workload survey in three conditions
for the simulation (\emph{left}) and in two conditions for the experiment
(\emph{right}). Conditions are as follows: A1: with no virtual fixtures
(only the entry-sphere constraint), A2, B1: with shaft-shaft collision
avoidance, A3, B2: with both LVFs and TGC. Values near the center
indicate better results.}
\end{figure}

\textcolor{black}{The users were asked to complete the NASA TLX workload
survey \cite{Hart2006} to evaluate the workload in terms of six indicators:}\textcolor{black}{\emph{
mental demand}}\textcolor{black}{, }\textcolor{black}{\emph{physical
demand, temporal demand, performance, effort}}\textcolor{black}{,
and }\textcolor{black}{\emph{frustration}}\textcolor{black}{. The
results of the survey are shown in Fig.~\ref{fig:ex1_NASA TLX}.
We normalized the scores of the NASA TLX to compare the relative scores
between techniques. The sum of the NASA TLX indicators for each condition
were 80.1 (for A1), 56.6 (for A2), 46.73 (for A3) out of 126. }

\textcolor{black}{Comparing A1 with A2, A1 had the highest workload
when summing up all indicators (p < 0.05). In special, A1 had a highest
score in }\textcolor{black}{\emph{mental demand}}\textcolor{black}{{}
(p < 0.05 ), }\textcolor{black}{\emph{performance}}\textcolor{black}{{}
(p < 0.05), }\textcolor{black}{\emph{frustration}}\textcolor{black}{{}
(p < 0.05), and }\textcolor{black}{\emph{effort}}\textcolor{black}{{}
(p < 0.05). There was no statistical significance in }\textcolor{black}{\emph{physical
demand}}\textcolor{black}{{} (p = 0.84) and }\textcolor{black}{\emph{temporal
demand}}\textcolor{black}{{} (p = 0.69). From the feedback we received
from the users, operating the system without the automatic constraints
increased their overall workload because they had to prevent collisions
while performing the looping task manually.}

\textcolor{black}{Comparing A2 with A3, A3 had the lowest sum of workload
indicators (p < 0.05). A3 significantly reduced the }\textcolor{black}{\emph{physical
demand}}\textcolor{black}{{} (p < 0.05) and }\textcolor{black}{\emph{temporal
demand}}\textcolor{black}{{} (p < 0.05) during the looping motion. There
was no significant difference in }\textcolor{black}{\emph{mental demand
}}\textcolor{black}{(p = 0.07)}\textcolor{black}{\emph{, performance
}}\textcolor{black}{(p = 0.06)}\textcolor{black}{\emph{, effort}}\textcolor{black}{{}
(p = 0.06), and }\textcolor{black}{\emph{frustration}}\textcolor{black}{{}
(p = 0.18).}

\textcolor{black}{These results show that, in terms of workload, A3
was the superior technique overall. }

\textcolor{black}{In addition to the workload survey, the users were
asked ``What was your biggest difficulty with the system/what aspects
of the system could be improved?'' as an open-ended question. Two
comments regarded the tunning of the system ``the motion scaling
is too small'' and ``the tools move too fast,'' which might be
addressed by a user-specific system tunning in future works. In addition,
some difficulties were specific to the task itself ``depth was hard
to perceive'' because of the 2D images, and ``the RCM constraint
was the major cause of inconvenience.'' }

\subsection{Experiment}

The experiment was designed to study the effect of the proposed methods
on the medical doctors during a robot-aided surgical looping. The
experimental setup replicated the simulation, and the experimental
parameters are shown in Table~\ref{tab:all-parameters}. A surgical
thread (5-0 PERMA-HAND SILK, ETHICON, USA) was held by the robot controlled
by the operator's dominant hand. In this experiment, the base of the
robots and the tooltip position with respect to the robotic end-effector
were calibrated by using a visual-tracking system (Polaris Spectra,
NDI, Canada) through a pivoting process \cite{Marinho2019}.

Two pediatric surgeons, one expert level (EL) and one intermediate
level (IL), were recruited. The surgeons were instructed to perform
the double loop under two conditions: B1 only with shaft\textendash shaft
collision avoidance and B2 with both LVFs and TGC. The double loop
was performed 10 times, 5 times for each condition. The users operated
the robots using the haptic interfaces, and the images captured by
the endoscope (Endoarm, Olympus, Japan) were displayed on a monitor.
All the conditions were assigned in random a sequence to reduce possible
biases.

\subsubsection{Results and discussion}

\paragraph{Controller performance}

\textcolor{black}{The medical doctors successfully conducted the surgical
looping under both conditions, }\textcolor{black}{\emph{i.e.,}}\textcolor{black}{{}
they looped the surgical thread twice about the left instrument, and
those loops were stable. In addition, no collisions happened.} The
task completion time and the error between the tooltip of $R_{2}$
and the surface of the guidance cylinder (Section\textcolor{black}{~}\ref{subsec:constrained_optimization})
were recorded during the experiment. The results are shown in Table.\textcolor{black}{~\ref{tab:all-time-shaftshaft}. }

\textcolor{black}{For the median task completion time for conditions
B1 and B2, there was no significant difference for the EL (p = 0.875)
and the IL (p = 0.32). The median task completion time for the IL
appears skewed towards a reduction of task completion time when using
the guidance virtual fixtures in condition B2.}

\textcolor{black}{The error between the $R_{2}$ tooltip and the guidance
cylinder decreased by an average of 33\% for the EL and by 11\% for
the IL. These results indicate that the virtual fixtures decreased
the error more for the EL than for the IL. We conjecture that the
more substantial reduction of error for the EL might be related to
how the EL operated the master interface. The EL moved the master
interfaces more softly than the IL, giving more room for the virtual
fixtures to assist and decrease the error. }

\textcolor{black}{These results indicate benefits for medical doctors
as well, but more users will be required in follow-up studies to investigate
this hypothesis further. }

\begin{table}[h]
\caption{\label{tab:all-time-shaftshaft} User performance in the experiment.}

\centering

\noindent\resizebox{0.8\columnwidth}{!}{%
\begin{centering}
\textcolor{black}{}%
\begin{tabular}{ccccccc}
\hline 
\noalign{\vskip\doublerulesep}
 & \multicolumn{2}{c}{Expert} & \multicolumn{2}{c}{Intermediate} & \multicolumn{2}{c}{Total}\tabularnewline[\doublerulesep]
\noalign{\vskip\doublerulesep}
\hline 
\noalign{\vskip\doublerulesep}
 & B1 & B2 & B1 & B2 & B1 & B2\tabularnewline[\doublerulesep]
\hline 
\noalign{\vskip\doublerulesep}
\noalign{\vskip\doublerulesep}
Median time {[}s{]} & 33 & 32 & 31.5 & 28 & 32 & 31\tabularnewline[\doublerulesep]
\noalign{\vskip\doublerulesep}
\noalign{\vskip\doublerulesep}
Mean error {[}mm{]} & 0.89 & 0.59 & 2.16 & 1.92 & 1.82 & 1.40\tabularnewline[\doublerulesep]
\noalign{\vskip\doublerulesep}
\end{tabular}
\par\end{centering}
}
\end{table}

\paragraph{User evaluation}

\textcolor{black}{The surgeons completed the same modified NASA TLX
survey, and there was no significant difference (p > 0.25) in the
workload between B1 and B2 (Fig.~\ref{fig:ex1_NASA TLX}). }These
results indicate that the proposed method successfully guided the
instruments closer to the desired surface without a noticeable difference
in the workload felt by the surgeon. 

\textcolor{black}{In the experimental study, we could observe a collision-free
looping with decreased overall looping error. This assistance is essential
when the medical doctors are learning how to use the robotic system
(even if they are experienced in manual surgery), to prevent damage
to the instruments and the anatomical model. A more comprehensive
experimental study with a larger number of surgical doctors is required
in follow-up studies to investigate at which point in their learning
curve the medical doctors can most benefit from the proposed assistance
method. }

\section{Conclusions}

\textcolor{black}{In this paper, two virtual fixtures for looping
were proposed, the looping virtual fixtures (LVFs) and the trajectory
guidance cylinder (TGC). These methods can improve the safety and
precision of the looping task in surgical suturing under the constrained
workspace of pediatric surgery. On the slave side, a constrained optimization
algorithm generates the LVFs and the TGC. On the master side, a Cartesian
impedance algorithm allows the user to ``feel'' safe directions
(LVFs) and optimal directions (TGC) during the looping. }The proposed
algorithm is evaluated in a simulation and an experiment with users
that show the safety and increased precision of the looping. \textcolor{black}{The
results of our simulation study indicate that the proposed methods
might be more effective when used by operators with less experience.}

\balance

\bibliographystyle{IEEEtran}
\bibliography{bib/ral}

\begin{thebibliography}{10}
\providecommand{\url}[1]{#1}
\csname url@samestyle\endcsname
\providecommand{\newblock}{\relax}
\providecommand{\bibinfo}[2]{#2}
\providecommand{\BIBentrySTDinterwordspacing}{\spaceskip=0pt\relax}
\providecommand{\BIBentryALTinterwordstretchfactor}{4}
\providecommand{\BIBentryALTinterwordspacing}{\spaceskip=\fontdimen2\font plus
\BIBentryALTinterwordstretchfactor\fontdimen3\font minus
  \fontdimen4\font\relax}
\providecommand{\BIBforeignlanguage}[2]{{%
\expandafter\ifx\csname l@#1\endcsname\relax
\typeout{** WARNING: IEEEtran.bst: No hyphenation pattern has been}%
\typeout{** loaded for the language `#1'. Using the pattern for}%
\typeout{** the default language instead.}%
\else
\language=\csname l@#1\endcsname
\fi
#2}}
\providecommand{\BIBdecl}{\relax}
\BIBdecl

\bibitem{Cundy2015a}
T.~P. Cundy, H.~J. Marcus, A.~Hughes-Hallett, T.~MacKinnon, A.~S. Najmaldin,
  G.-Z. Yang, and A.~Darzi, ``Robotic versus non-robotic instruments in
  spatially constrained operating workspaces: a pre-clinical randomized
  crossover study,'' \emph{{BJU} International}, vol. 116, no.~3, pp. 415--422,
  may 2015.

\bibitem{Takazawa2018}
S.~Takazawa, T.~Ishimaru, K.~Harada, K.~Deie, A.~Hinoki, H.~Uchida, N.~Sugita,
  M.~Mitsuishi, T.~Iwanaka, and J.~Fujishiro, ``Evaluation of surgical devices
  using an artificial pediatric thoracic model: A comparison between
  robot-assisted thoracoscopic suturing versus conventional video-assisted
  thoracoscopic suturing,'' \emph{Journal of Laparoendoscopic {\&} Advanced
  Surgical Techniques}, vol.~28, no.~5, pp. 622--627, may 2018.

\bibitem{marinho2019integration}
M.~M. Marinho, K.~Harada, A.~Morita, and M.~Mitsuishi, ``Smartarm: Integration
  and validation of a versatile surgical robotic system for constrained
  workspaces,'' \emph{The International Journal of Medical Robotics and
  Computer Assisted Surgery (IJMRCAS)}, 2019, (in press).

\bibitem{Arata2019}
J.~Arata, Y.~Fujisawa, R.~Nakadate, K.~Kiguchi, K.~Harada, M.~Mitsuishi, and
  M.~Hashizume, ``Compliant four degree-of-freedom manipulator with locally
  deformable elastic elements for minimally invasive surgery,'' in \emph{2019
  International Conference on Robotics and Automation ({ICRA})}.\hskip 1em plus
  0.5em minus 0.4em\relax {IEEE}, May 2019.

\bibitem{Marinho2019a}
M.~M. Marinho, B.~V. Adorno, K.~Harada, K.~Deie, A.~Deguet, P.~Kazanzides,
  R.~H. Taylor, and M.~Mitsuishi, ``A unified framework for the teleoperation
  of surgical robots in constrained workspaces,'' in \emph{2019 International
  Conference on Robotics and Automation ({ICRA})}.\hskip 1em plus 0.5em minus
  0.4em\relax {IEEE}, may 2019.

\bibitem{Cundy2015}
T.~P. Cundy, H.~J. Marcus, A.~Hughes-Hallett, S.~Khurana, and A.~Darzi,
  ``Robotic surgery in children: adopt now, await, or dismiss?''
  \emph{Pediatric Surgery International}, vol.~31, no.~12, pp. 1119--1125, Sep.
  2015.

\bibitem{Osa2018}
T.~Osa, N.~Sugita, and M.~Mitsuishi, ``Online trajectory planning and force
  control for automation of surgical tasks,'' \emph{{IEEE} Transactions on
  Automation Science and Engineering}, vol.~15, no.~2, pp. 675--691, Apr. 2018.

\bibitem{kapoor2005spatial}
A.~Kapoor, M.~Li, and R.~H. Taylor, ``Spatial motion constraints for robot
  assisted suturing using virtual fixtures,'' in \emph{International Conference
  on Medical Image Computing and Computer-Assisted Intervention}.\hskip 1em
  plus 0.5em minus 0.4em\relax Springer, May 2005, pp. 89--96.

\bibitem{kapoor2006constrained}
------, ``Constrained control for surgical assistant robots.'' in \emph{2006
  International Conference on Robotics and Automation ({ICRA})}, May 2006, pp.
  231--236.

\bibitem{Sen2016}
S.~Sen, A.~Garg, D.~V. Gealy, S.~McKinley, Y.~Jen, and K.~Goldberg,
  ``Automating multi-throw multilateral surgical suturing with a mechanical
  needle guide and sequential convex optimization,'' in \emph{2016 {IEEE}
  International Conference on Robotics and Automation ({ICRA})}.\hskip 1em plus
  0.5em minus 0.4em\relax {IEEE}, May 2016.

\bibitem{chen2016virtual}
Z.~Chen, A.~Malpani, P.~Chalasani, A.~Deguet, S.~S. Vedula, P.~Kazanzides, and
  R.~H. Taylor, ``Virtual fixture assistance for needle passing and knot
  tying,'' in \emph{IEEE/RSJ International Conference on Intelligent Robots and
  Systems (IROS)}.\hskip 1em plus 0.5em minus 0.4em\relax IEEE, Oct. 2016, pp.
  2343--2350.

\bibitem{Fontanelli2018}
G.~A. Fontanelli, G.-Z. Yang, and B.~Siciliano, ``A comparison of assistive
  methods for suturing in {MIRS},'' in \emph{2018 {IEEE}/{RSJ} International
  Conference on Intelligent Robots and Systems ({IROS})}.\hskip 1em plus 0.5em
  minus 0.4em\relax {IEEE}, oct 2018.

\bibitem{bowyer2014active}
S.~A. Bowyer, B.~L. Davies, and F.~R. y~Baena, ``Active constraints/virtual
  fixtures: A survey,'' \emph{IEEE Transactions on Robotics}, vol.~30, no.~1,
  pp. 138--157, 2014.

\bibitem{Kapoor2008}
A.~Kapoor and R.~H. Taylor, ``A constrained optimization approach to virtual
  fixtures for multi-handed tasks,'' in \emph{2008 International Conference on
  Robotics and Automation ({ICRA})}.\hskip 1em plus 0.5em minus 0.4em\relax
  {IEEE}, may 2008.

\bibitem{Selvaggio2019}
M.~Selvaggio, A.~Ghalamzan, R.~Moccia, F.~Ficuciello, and B.~Siciliano,
  ``Haptic-guided shared control for needle grasping optimization in minimally
  invasive robotic surgery,'' in \emph{2019 {IEEE}/{RSJ} International
  Conference on Intelligent Robots and Systems ({IROS})}, Nov. 2019.

\bibitem{Banach2019}
A.~Banach, K.~Leibrandt, M.~Grammatikopoulou, and G.-Z. Yang, ``Active
  contraints for tool-shaft collision avoidance in minimally invasive
  surgery,'' in \emph{2019 International Conference on Robotics and Automation
  ({ICRA})}.\hskip 1em plus 0.5em minus 0.4em\relax {IEEE}, May 2019.

\bibitem{Looi2013}
T.~Looi, B.~Yeung, M.~Umasthan, and J.~Drake, ``{KidsArm} an image-guided
  pediatric anastomosis robot,'' in \emph{2013 {IEEE}/{RSJ} International
  Conference on Intelligent Robots and Systems}.\hskip 1em plus 0.5em minus
  0.4em\relax {IEEE}, Nov. 2013.

\bibitem{Marinho2019}
M.~M. Marinho, B.~V. Adorno, K.~Harada, and M.~Mitsuishi, ``Dynamic active
  constraints for surgical robots using vector-field inequalities,''
  \emph{{IEEE} Transactions on Robotics}, vol.~35, no.~5, pp. 1166--1185, oct
  2019.

\bibitem{cheng1994}
F.-T. Cheng, T.-H. Chen, and Y.-Y. Sun, ``Resolving manipulator redundancy
  under inequality constraints,'' \emph{IEEE Transactions on Robotics and
  Automation}, vol.~10, no.~1, pp. 65--71, Feb 1994.

\bibitem{bowyer2013dynamic}
S.~A. Bowyer and F.~R. y~Baena, ``Dynamic frictional constraints for robot
  assisted surgery,'' in \emph{World Haptics Conference (WHC), 2013}.\hskip 1em
  plus 0.5em minus 0.4em\relax IEEE, 2013, pp. 319--324.

\bibitem{Harada2016}
K.~Harada, G.~Ishikawa, S.~Takazawa, T.~Ishimaru, N.~Sugita, T.~Iwanaka, and
  M.~Mitsuishi, ``Development of a neonatal thoracic cavity model and
  preliminary study,'' \emph{Journal of Japan Society of Computer Aided
  Surgery}, vol.~18, no.~2, pp. 80--86, 2016.

\bibitem{Kim2018}
S.~Kim, J.~L. Buendia, S.~A. Heredia-Perez, M.~M. Marinho, K.~Harada,
  N.~Kaneko, T.~Ushiku, and M.~Mitsuishi, ``Towards the automation of grossing
  in pathology examinations using industrial robotic arms,'' in \emph{The 14th
  Asian Conference on Computer Aided Surgery (ACCAS)}, Nov. 2018, pp. 202--203.

\bibitem{williams1949experimental}
E.~Williams, ``Experimental designs balanced for the estimation of residual
  effects of treatments,'' \emph{Australian Journal of Chemistry}, vol.~2,
  no.~2, pp. 149--168, 1949.

\bibitem{Hart2006}
S.~G. Hart, ``Nasa-task load index ({NASA}-{TLX}) 20 years later,''
  \emph{Proceedings of the Human Factors and Ergonomics Society Annual
  Meeting}, vol.~50, no.~9, pp. 904--908, oct 2006.

\end{thebibliography}

\end{document}